\documentclass[twoside,leqno,twocolumn]{article}

\usepackage[letterpaper]{geometry}

\usepackage{ltexpprt}
\usepackage{algorithm}
\usepackage{algorithmic}

\usepackage{amsmath,amssymb,amsfonts}
\usepackage{graphicx}
\usepackage{textcomp}
\usepackage{xcolor}
\usepackage{booktabs} 
\usepackage{amsmath}
\usepackage{graphicx}
\usepackage{array, booktabs}
\usepackage{subfigure}
\usepackage{cite}
\usepackage{multirow}
\usepackage{amsfonts,bm}
\usepackage{braket}
\usepackage{mathtools}
\usepackage{enumerate}
\usepackage{enumitem}
\usepackage{empheq}
\usepackage{calc}
\usepackage{balance}
\usepackage{dsfont}
\usepackage{textcomp} 
\usepackage{color, colortbl}
\usepackage{caption}

\begin{document}

\title{\Large Graph-based Reinforcement Learning for  Active Learning in Real Time: \\An Application in Modeling River Networks}%\thanks{Supported by GSF grants ABC123, DEF456, and GHI789.}}
% \author{Corey Gray\thanks{Society for Industrial and Applied Mathematics.}
% \and Tricia Manning\thanks{Society for Industrial and Applied Mathematics.}}

\author{Xiaowei Jia$^{1}$,  Beiyu Lin$^2$, Jacob Zwart$^3$, Jeffrey Sadler$^3$, Alison Appling$^3$, \\Samantha Oliver$^3$, Jordan Read$^3$\\
\small\baselineskip=9pt $^1$ University of Pittsburgh, 
\small\baselineskip=9pt $^2$Univesity of Texas - Rio Grande Valley, 
\small\baselineskip=9pt $^3$ U.S. Geological Survey\\
\small  $^1$xiaowei@pitt.edu,$^2$beiyu.lin@utrgv.edu,$^3$\{jzwart,jsadler,aappling,soliver,jread\}@usgs.gov
}

\date{}

\maketitle

% Copyright Statement
% When submitting your final paper to a SIAM proceedings, it is requested that you include 
% the appropriate copyright in the footer of the paper.  The copyright added should be 
% consistent with the copyright selected on the copyright form submitted with the paper.
% Please note that "20XX" should be changed to the year of the meeting.

% Default Copyright Statement
\fancyfoot[R]{\scriptsize{Copyright \textcopyright\ 20XX by SIAM\\
Unauthorized reproduction of this article is prohibited}}

% Depending on which copyright you agree to when you sign the copyright form, the copyright 
% can be changed to one of the following after commenting out the default copyright statement
% above.

%\fancyfoot[R]{\scriptsize{Copyright \textcopyright\ 20XX\\
%Copyright for this paper is retained by authors}}

%\fancyfoot[R]{\scriptsize{Copyright \textcopyright\ 20XX\\
%Copyright retained by principal author's organization}}

%\pagenumbering{arabic}
%\setcounter{page}{1}%Leave this line commented out.

\begin{abstract} \small\baselineskip=9pt 
%Given the success of machine learning (ML), especially deep learning, on commercial applications, there is a growing interest in using advanced ML models  to accelerate scientific discovery. However, e
Effective training of advanced ML models requires large amounts of labeled data, which is often scarce in scientific problems given the substantial human labor and material cost to collect labeled data. This poses a challenge on determining when and where we should deploy measuring instruments (e.g., in-situ sensors) to collect labeled data efficiently. This problem differs from traditional pool-based active learning settings in that  the labeling decisions have to be made immediately after we observe the input data that come in a time series. In this paper, we develop a real-time active learning method that uses the spatial and temporal contextual information to select representative query samples in a reinforcement learning framework. %to determine the representatitveness of each sample in the selection process. 
To reduce the need for large training data, we further propose to transfer the policy learned from  simulation data which is generated by existing physics-based models. We demonstrate the effectiveness of the proposed method by predicting streamflow and water temperature in the Delaware River Basin given a limited  budget for collecting labeled data. We further study the spatial and temporal distribution of selected samples to verify the ability of this method in selecting informative samples over space and time.  
\end{abstract}

\section{Introduction}
The last few years have witnessed a surge of interest in building machine learning (ML) methods  for scientific applications in diverse disciplines, e.g., hydrology \cite{jia2019physics}, biological sciences \cite{yazdani2019systems}, %earth systems \cite{reichstein2019deep}, 
and climate science~\cite{faghmous2014big}.  %\cite{krasnopolsky2006complex,faghmous2014big,o2018using}.  %turbulence modeling \cite{mohan2018deep,bode2019using,xiao2019reduced}, material discovery \cite{raccuglia2016machine,cang2018improving,schleder2019dft}, and quantum chemistry \cite{sadowski2016synergies,schutt2017schnet}). 
Given the promising results from previous research,  expectations are rising for using ML to accelerate scientific discovery and help address some of the biggest challenges that are facing humanity such as water quality, climate, and healthcare. However, ML models focus on mining the statistical relationships from data and thus often require  large amount of labeled observation data to tune their model parameters.

\begin{figure} [!t]
\centering
\subfigure[]{ \label{fig:b}{}
\includegraphics[width=0.44\columnwidth]{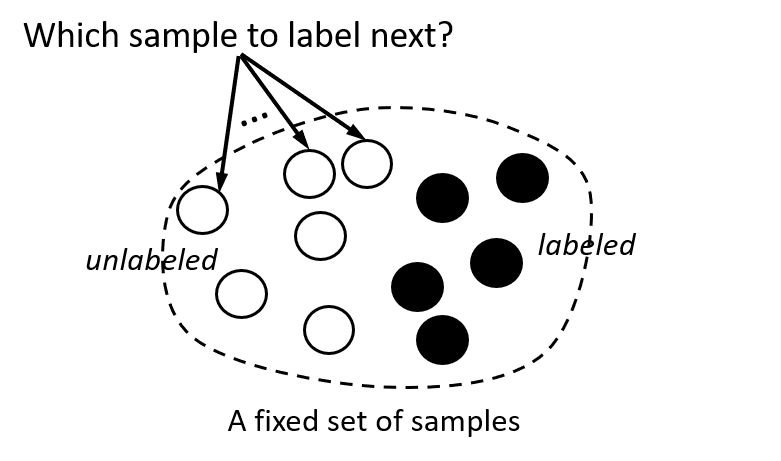}
}\hspace{-.3in}
\subfigure[]{ \label{fig:b}{}
\includegraphics[width=0.54\columnwidth]{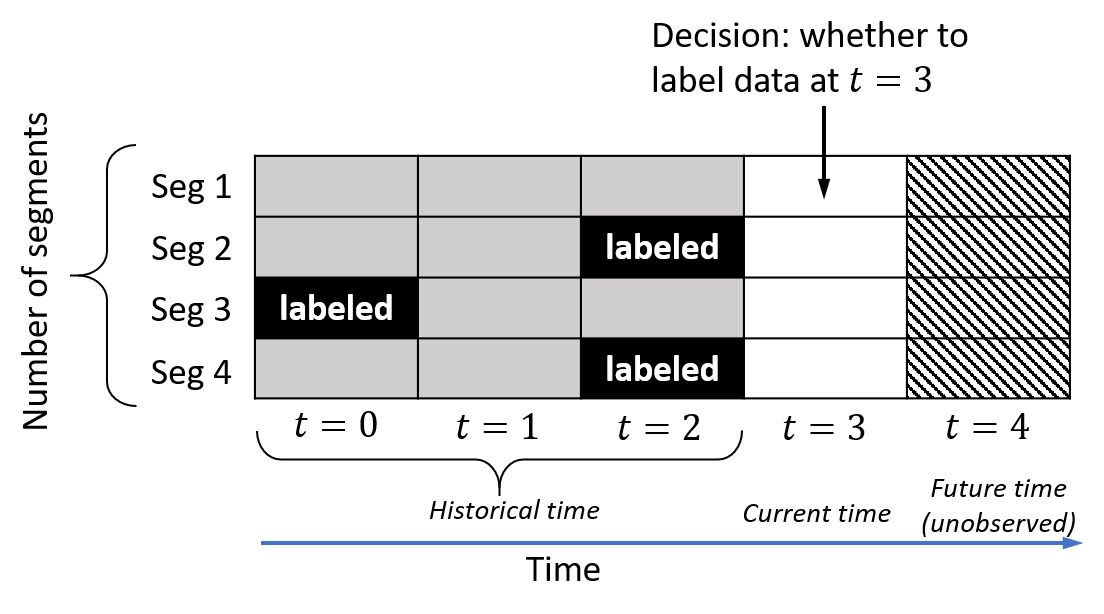}
}
\vspace{-.1in}
\caption{(a) Pool-based active learning. (b) Real-time active learning.}
\label{fig:AL}
%\vspace{-.3in}
\end{figure}

Collecting labeled data is often expensive in scientific applications due to the substantial manual labor and material cost required to deploy sensors or other measuring instruments.  %For example, collecting water temperature data commonly require people with hydrologic expertise to sail a boat to the middle of a lake or river and deploy sensors at target locations. 
For example, collecting water temperature data commonly requires highly trained scientists to travel to sampling locations and deploy sensors within a lake or stream, incurring personnel and equipment costs for data that may not improve model predictions. 
To  make the data collection more efficient, we aim to develop a data-driven method that assists  domain experts in determining  when and where to deploy measuring instruments in real time so that data collection can be optimized for training ML models. %we can  collect most representative samples  for training ML models. 
%To  make the data collection more efficient, we aim to develop a data-driven method to actively select query samples in real time. Such a model will assist the domain experts in determining  when and where to deploying measuring instruments so that we can  collect most representative samples  for training ML models. %new labeled samples. 

%Active learning techniques provide a data-driven solution to select 
Active learning has shown great promise for selecting representative samples~\cite{settles2009active,felder2009active}. In particular, it aims to find a query strategy based on which %that select most representative samples for annotation with the hope that the collected labeled samples can 
we can annotate samples that optimize the  training of a predictive ML model. Traditional pool-based active learning methods are focused on selecting query samples 
from a fixed set of data points (Fig.~\ref{fig:AL} (a)). %and thus reduce the cost of data collection. These techniques 
These techniques have been widely explored in image recognition~\cite{li2013adaptive,gal2017deep} %\cite{li2013adaptive,vijayanarasimhan2014large,gal2017deep} 
and natural language processing~\cite{shen2017deep,zhou2013active}. %, they are focused on query samples from  
%, i.e., when and where should we collect new labeled samples. 
%Most traditional active learning methods  
%and they can be used to 
%for selecting new query samples. 
%In each round, the model finds the sample for which the current ML model 
These approaches mostly select samples based on their uncertainty level, which can be measured by Bayesian inference~\cite{kendall2017uncertainties} and Monte Carlo drop-out approximation~\cite{gal2016dropout}. 
% lowest confidence for generating the predictions. 
The samples with higher uncertainty tend to stay closer to the current decision boundary %of the current model 
and thus can bring higher information gain to refine the boundary. Some other approaches also explore the diversity of samples so that they can annotate samples different with those that have already been labeled~\cite{sener2017active,cai2017active,wu2019active}.

% Most traditional active learning methods  select new query samples from a fixed set of unlabeled data points based on their prediction uncertainty level. In each round, the model finds the sample for which the current ML model has lowest confidence for generating the predictions. The intuition is that the samples with higher uncertainty are close to the decision boundary of the current model and thus can be used to better refine the classification boundary.  The uncertainty level can be commonly measured using Bayesian inference~\cite{kendall2017uncertainties} and drop-out approximation~\cite{gal2016dropout}. %\textcolor{red}{Add citations here}

However, the pool-based active learning approaches cannot be used in scientific problems as they assume all the data points are available in  a fixed set %, and thus they cannot  
%cannot be used 
while scientific data have to be annotated in real time. When monitoring scientific systems, new labels can only be collected by deploying sensors or  other measuring instruments. Such labeling decisions have to be made immediately %after we make decisions and we cannot change decisions (labeling or not) that we have made on past events.  
%The decisions (i.e., label or not) have to be made immediately  
after we observe the data at the current time, which requires balancing the information gain against the budget cost. Hence, these labeling decisions are made without access to future data and also cannot be changed afterwards. Such a real-time labeling process (also referred to as stream-based selective sampling) is described in Fig.~\ref{fig:AL}. %The decision has, we cannot change it in the future.  %Besides, we cannot change decisions that have been made on past events.  
%can deploy sensors in real time. 
%In scientific applications, we cannot foresee future data points or change decisions (deploy sensors or not) we made on past events.
%Besides, we have to consider whether we should spend limited budget to label currently observed samples.

Moreover, existing approaches do not take into account the representativeness of the selected samples given their spatial and temporal context. Scientific systems commonly involve multiple physical processes that evolve over time and also interact with each other. For example, in a river network, different river segments  can have different thermodynamic patterns due to different catchment characteristics as well as climate conditions. Connected river segments can also interact with each other through the water advected from upstream to downstream segments.  In this case,  new annotated samples can be less helpful if they are selected from river segments that have similar spatio-temporal patterns with previously labeled samples. Instead, the model should take samples that cover different time periods and river segments with distinct properties. % (e.g., headwater, high-flow river segments and low-flow segments in a river network). 

% However, these approaches do not take into account the representativeness of the selected samples given the spatial and temporal context each process. For example, the new query samples are less helpful if they have similar spatio-temporal patterns with previously selected samples. Instead, the model should take samples that cover different time periods and different processes with distinct properties (e.g., headwater, high-flow river segments and low-flow segments in a river network). 

% Another major limitation is that these approaches can only select query samples from a fixed set of data points from a retrospective perspective, but cannot be used to determine whether we can deploy sensors in real time. In scientific applications, we cannot foresee future data points or change decisions (deploy sensors or not) we made on past events. Hence, we need to better estimate whether we should spend limited budget to collect labels on the currently observed samples. 

To address these challenges, we propose a new framework \textbf{G}raph-based \textbf{R}einforcement Learning for  \textbf{Rea}l-Time \textbf{L}abeling (GR-REAL), in which we formulate real-time active learning problem as a Markov decision process. 
This proposed framework is developed in the context of modeling streamflow and water temperature in river networks but the framework can be generally applied to many complex physical systems with interacting processes.  Our method makes  labeling decisions based on the spatial and temporal context of each river segment as well as the uncertainty level  at the current time.  %on this segment as its current state. 
Once we determine the actions of whether to label each segment at the current time step, the collected labels can be used to refine %We aim to automatically decide whether we need to label each segment at the current time step and the collected labels can be used to refine 
the way we represent the spatio-temporal context and estimate uncertainty for the observed samples at the next time step (i.e., the next state).  %help generate predictions for the next time step or the next state.  % of target variables. 

%formulate the active labeling problem as a Markov decision making process to dynamically determine whether we will label the currently observed samples from different objects. 
%In particular, %we build two models, 

In particular, the proposed framework consists of a predictive model and a decision model. The predictive model extracts spatial and temporal dependencies from data and embeds such contextual information in a state vector. The predictive model also generates final predictions and estimates uncertainty based on the obtained embeddings. At the same time, the decision model is responsible for determining whether we will take labeling actions at the current time step based on the embeddings and outputs obtained from the predictive model. The collected labels are then used to refine the predictive model.    %can be the DRGrN model (proposed in Section~\ref{sec:rgrn}) that can extract the spatial and temporal patterns for each sample.
% In particular, the decision model %combines embeddings extracted by the predictive model, the model uncertainty, the remaining budget 
% utilizes the spatio-temporal patterns embedded in the hidden representation of DRGrN model (Section~\ref{sec:rgrn}) and model uncertainty levels to predict the potential reward of labeling or not labeling the current data point. 
% Such a reward can be estimated as the expectation of accumulated performance improvement via dynamic programming over training sequential data~\cite{mnih2013playing}. We will iteratively update the %predictive 
% DRGrN model and the decision model. In each round, the selected new samples by the decision model will be used to update the DRGrN model. Then the hidden representation and the confidence resulted from the DRGrN model will be used to update the potential reward for training the decision model. 
%To address these challenges, 
We train the decision model via reinforcement learning using past observation data. % to make labeling decisions. %algorithm to dynamically determine whether we need to label the currently observed samples from different processes. 
%In particular, we construct two models, predictive model and decision model. The predictive model can be the RGrN model (proposed in Section~\ref{sec:rgrn}) that can extract the spatial and temporal patterns for each sample. 
%The decision model combines embeddings extracted by the predictive model, the model uncertainty, the remaining budget to 
%predict the potential 
During the training phase, the reward of labeling each river segment at each time step %current time. During the training process, such reward 
can be estimated as the expectation of accumulated performance improvement via dynamic programming over training sequential data. %In each round, the selected new samples by the decision model will be used to update the predictive model. 

Since this proposed data-driven method requires separate training data from the past history, which can be scarce in many scientific systems, we also propose a way to transfer knowledge from existing physics-based models which are commonly used by domain scientists to study environmental and engineering problems. The transferred knowledge can be used to initialize the decision model and  thus  less training data is required to fine-tune it  to a quality model.  %Specifically, we will use the dense simulations generated by physics-based models to initialize both the predictive model and the decision model. 

We evaluate our proposed framework in predicting streamflow and water temperature in the Delaware River Basin. Our proposed method produces superior prediction performance given limited budget for labeling. We also show that  the distribution of collected samples is consistent with the dynamic patterns in river networks.

\section{Related Work}

% active learning
Active learning techniques have been used to intelligently select query samples that help improve the learning performance of ML models~\cite{settles2009active}. These methods have shown much success in annotating image data~\cite{li2013adaptive,gal2017deep} %\cite{gal2017deep,wang2016cost,yang2017suggestive} 
and text data~\cite{shen2017deep,zhou2013active}. In those problems, we can always hire human experts to visually annotate samples at any time after data have been collected because the mapping  from input data (e.g., images, text sentences or short phrases) to labels have been perceived by human. The major difference in scientific problems is that the relationships from data samples to labels cannot be fully captured by human but often requires  specific measuring instruments deployed by domain scientists. Hence, these measurement must be taken instantly after we observe the data and the decisions cannot be changed afterwards.

Such real-time active learning tasks are also referred to as stream-based selective sampling~\cite{settles2009active}. Traditional stream-based selective sampling approaches rely on heuristic or statistical metrics to select informative data samples~\cite{dagan1995committee,smailovic2014stream,cohn1996active}  %~\cite{dagan1995committee,smailovic2014stream,cohn1996active} 
and cannot fully exploit the complex relationships between data samples in a long sequence and estimate the long-term potential reward of labeling each sample. More recently, reinforcement learning has been used to learn how-to-label behaviors~\cite{cheng2013feedback,wassermann2019ral}. However, these methods do not consider the spatial and temporal context of each sample, which is often important for determining the information gain of labeling the data. Besides, they commonly require a separate large training set, which is hard to obtain in scientific applications.

% active learning+reinforcement learning

% Graph NN
Recent advances in deep learning models have brought a huge potential for representing spatial and temporal context. For example, the Long-Short Term Memory (LSTM) has found great success in capturing temporal dependencies~\cite{jia2019physics} in scientific problems.  
Also, the Graph Convolutional Networks (GCN) model %has been widely used in commercial applications and it 
has also proven to be effective in %automatically extracting latent factors that influence the neighbors in a graph. 
representing interactions between multiple objects. 
Given its unique capacity,  GCN has achieved the improved prediction accuracy in  several scientific problems~\cite{qi2019hybrid,xie2018crystal}. %For example, Moshe et al.~\cite{moshe2020hydronets_backup} propose HydroNets,  which combines the information from each river segment and its upstream segments for improving streamflow predictions. It also learns local patterns for each basin by introducing basin-specific model layers in addition to the global model. %This method focuses on predicting basins that are well monitored and it remains limited in generalizing to different scenarios or learning with less data. In contrast, we leverage the prior physical knowledge to learn latent variables that make the model more generalizable. 

%However, the information propagated amongst nodes in GCN is essentially an  abstract representation learned by end-to-end training. 
%Such abstract representations are not meant to enforce consistency with known physical relationships among different processes, such as in river networks. 

% simulation data
Simulation data have been used to assist in training ML models~\cite{jia2019physics,read2019process,shah2018airsim}. Since many ML models require an initial choice of model parameters before training, researchers have explored different ways to physically inform a model starting state. %Poor initialization can cause models to anchor in local minima, which is especially true for deep neural networks. %\textit{Transfer learning} can effectively tackle this issue, where the pre-trained models from a related task are fine-tuned with limited training data to fit the desired task. 
One way to harness physics-based modeling knowledge is to use the physics-based model's simulated data to pre-train the ML model, which also alleviates data paucity issues. Jia \textit{et al.} extensively discuss this strategy~\cite{jia2019physics}. They pre-train their Physics-Guided Recurrent Neural Network (PGRNN) models for lake temperature modeling on simulated data generated from a physics-based model and fine-tune it with minimal observed data. They show that pre-training %, even using data from a physical model with an uncalibrated set of parameters, 
can %still 
significantly reduce the training data needed for a quality model. In addition, Read et al.~\cite{read2019process} show that such models are able to generalize better to unseen scenarios. % than pure physics-based models.

\section{Problem Definition and Preliminaries}

\subsection{Problem definition}
Our objective is to model the dynamics of temperature and streamflow in a set of connected river segments given limited budget for collecting labels. We represent the connections amongst these river segments in a graph structure $\mathcal{G} = \{\mathcal{V},\mathcal{E},\textbf{W}\}$, where $\mathcal{V}$ represents the set of $N$ river segments and $\mathcal{E}$ represents the set of connections amongst river segments. Specifically,  we create an edge $(i,j)\in\mathcal{E}$ if the segment $j$ is anywhere downstream of the segment $i$. 
The matrix $\textbf{W}$ represents the adjacency level between each pair of segments, i.e.,  $\textbf{W}_{ij}=0$ means there is no edge from the segment $i$ to the segment $j$ and a higher value of $\textbf{W}_{ij}$ indicates that the segment $i$ is closer to the segment $j$ in terms of the stream distance. 
More details of  the adjacency matrix are discussed in Section~\ref{sec:dataset}.

We are provided with a time series of inputs for each river segment at daily scale. The input features $\textbf{x}_{i}^t$ for each segment $i$ at time $t$ are a $D$-dimensional vector, which include meteorological drivers, geometric parameters of the segments, etc. (more details can be found in Section~\ref{sec:dataset}). Assume currently we are at time~$T$. Once we observe input features $\{\textbf{x}_{i}^{t}\}_{i=1}^N$ at each future time step $t=T+1$ to $T+M$ for all the segments, we have to determine immediately whether to collect labels $\textbf{y}^t_i$ for each segment $i$. We also have to ensure that the labeling cost will not exceed the budget limit. %$B$.

To train our proposed model, we assume that we have access to observation data in the history. In particular, for each segment $i$, we have input features $\textbf{X}_i=\{\textbf{x}_{i}^{1}, \textbf{x}_{i}^2, ..., \textbf{x}_{i}^T\}$ and their labels $\textbf{Y}=\{y^t_i\}$. As elaborated later in the method and result discussion, we also divide the time period from $t=1$ to $T$ into separate training and hold-out periods.   %These labels are only collected on certain dates and certain segments.  

% For each river segment $i$, we have access to its input features at multiple time steps $\textbf{X}_i=\{\textbf{x}_{i}^{1}, \textbf{x}_{i}^2, ..., \textbf{x}_{i}^T\}$.  The input features $\textbf{x}_{i}^t$ are a $D$-dimensional vector, which include meteorological drivers, geometric parameters of the segments, etc. (more details can be found in Section~\ref{sec:dataset}). We also have a set of observed target variables $\textbf{Y}=\{y^t_i\}$ but they are only available for  certain time steps $t\in \{1,...,T\}$ and certain segments $i\in\{1,...,N\}$.

\subsection{Physics-based Streamflow and Temperature Model}
% \yell{Jake: please help check this and add more details if necessary. }

The Precipitation-Runoff Modeling System (PRMS)~\cite{markstrom2015prms} and the coupled Stream Network Temperature Model (SNTemp)~\cite{sanders2017documentation} is a physics-based model that simulates daily streamflow and water temperature for river networks, as well as other variables. PRMS is a one-dimensional, distributed-parameter modeling system that translates spatially-explicit meteorological information into water information including evaporation, transpiration, runoff, infiltration, groundwater flow, and streamflow. PRMS has been used to simulate catchment hydrologic variables relevant to management decisions at regional~\cite{lafontaine2013application} to national scales~\cite{regan2018description}, among other applications. The SNTemp module for PRMS %was based on the standalone, physics-based stream temperature model developed by Theurer et al.~\cite{theurer1984instream}, and 
simulates mean daily stream water temperature for each river segment by solving an energy mass balance model which accounts for the effect of inflows (upstream, groundwater, surface runoff), outflows, and surface heating and cooling on heat transfer in each river segment. The SNTemp module is driven by the same meteorological drivers used in PRMS and also driven by the hydrologic information simulated by PRMS (e.g. streamflow, groundwater flow). %For our application, we use a cutout of the National Hydrologic Model application of PRMS for the Delaware River Basin (our modeling domain), which includes a national-scale geospatial modeling fabric and spatially-explicit parameterization of various PRMS parameters~\cite{regan2018description}. 
Calibration of PRMS-SNTemp is very time-consuming because it involves a large number of parameters (84 parameters) and the parameters interact with each other both within segments and across segments.

\section{Method}

%In this section, we start with the description of the   GR-REAL framework and  the decision model. Then we discuss the extraction of spatio-temporal patterns using the predictive model. Finally, we  introduce a policy transfer method that helps initialize the GR-REAL framework by leveraging the simulation data produced by physics-based models. 

\begin{figure} [!t]
\centering
\includegraphics[width=0.85\columnwidth]{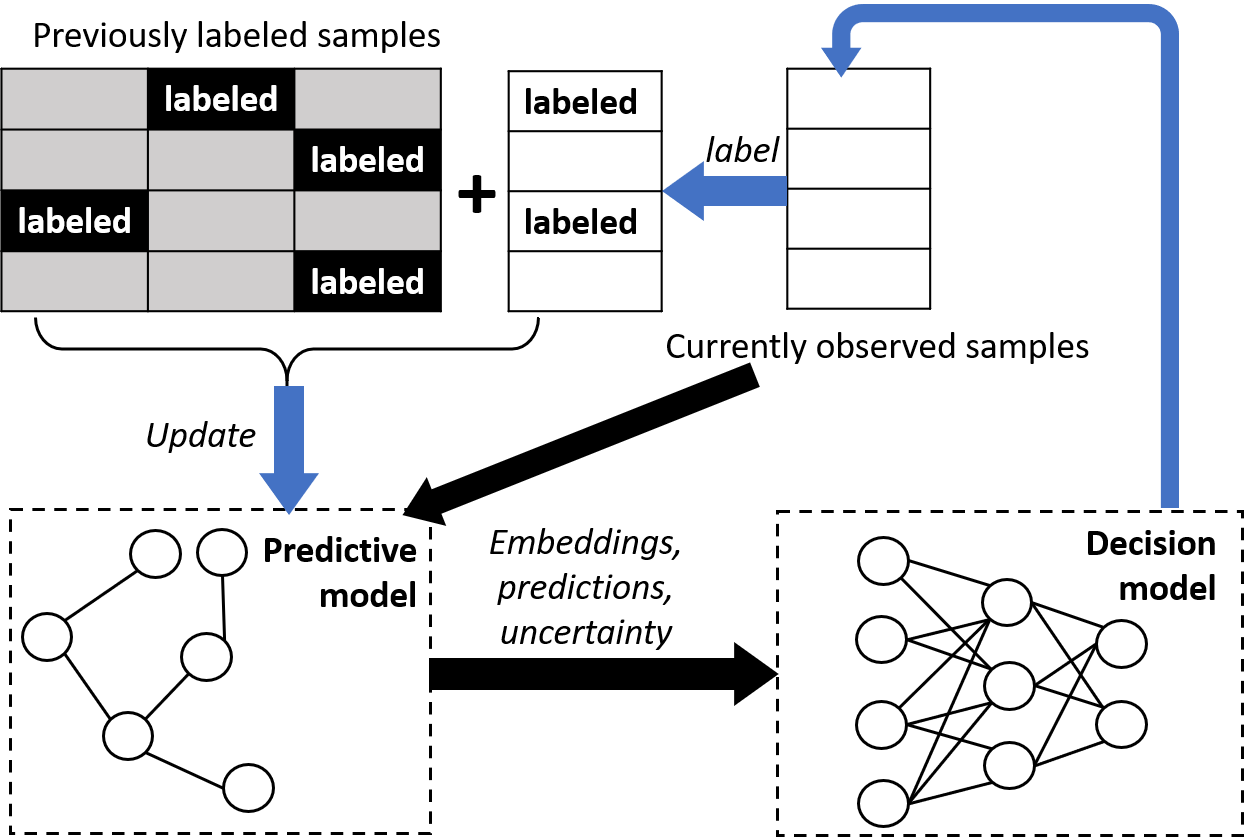}
\caption{The overview of GR-REAL framework. The black arrows shows the feed-forward process using the predictive model. The blue arrows shows the feedback process by the decision model.}
\label{fig:framework}
%\vspace{-.15in}
\end{figure}

%In this problem, we 
The proposed GR-REAL framework aims to actively train a predictive model in real time from streaming data,  %. We show the flow of the GR-REAL framework 
as shown in Fig.~\ref{fig:framework}. 
At each time, the predictive model embeds currently observed samples by incorporating the spatio-temporal context and provide its outputs (embeddings, predictions and uncertainty) to the decision model. Then the decision model determines to label a subset (or none) of observed samples at the current time. The labeled samples are then used to update the predictive model.  In the following, we will describe the predictive model and the decision model. % in details. 

\subsection{Recurrent Graph Neural Network}

%In a river network, most river segments are poorly observed or completely unobserved. To this end, we introduce a global ML model architecture, Recurrent Graph Network (RGrN), which is trained using data collected from all the river segments. 
Effective modeling of river segments requires the ability to capture their temporal thermodynamics and the influence received from upstream segments. Hence, we incorporate the information  from both previous time steps and neighbors (i.e., upstream segments) when modeling each segment. 
%Here we describe the recurrent process of generating the hidden representation $\textbf{h}^t$ from $\textbf{h}^{t-1}$, and we repeat this process for the entire sequence from $t=2$ to $T$ ($\textbf{h}^1$ learned from an LSTM model). 
Our model is based on the LSTM model which has proven to be effective in capturing long-term dependencies. 
Different from the standard LSTM, we develop a customized recurrent cell that combines the spatial and temporal context. 
%In the following discussion, we omit the subscript $i$ as we do not target a specific river segment.

When applied to each segment $i$ at time $t$, the recurrent cell has a cell state $\textbf{c}_i^t\in \mathbb{R}^H$, which serves as a memory and allows 
preserving the information from its history and neighborhood. Then the recurrent cell outputs a hidden representation $\textbf{h}_i^t\in \mathbb{R}^H$, from which we generate the target output. In the following, we describe the recurrent process of generating $\textbf{c}_i^t$ and $\textbf{h}_i^t$ based on the input $\textbf{x}_i^t$ and collected information from the previous time $t-1$.

For each river segment $i$ at time $t-1$, the model extracts latent variables %from this segment 
which contain relevant information to pass to its downriver segments. We refer to these latent variables as transferred variables. For example, the amount of water advected from each segment and its water temperature can directly impact the change of water temperature for its downriver segments.  %.   
%Formally, we generate transferred variables 
 %that contain the relevant information to pass to its downriver segments. The transferred variables are computed
%and they are generated 
We generate the transferred variables $\textbf{q}_{i}^{t-1}$ from the hidden representation $\textbf{h}_{i}^{t-1}$ at the previous time step: %, as follows: %follows:
\begin{equation}
\small
    \textbf{q}_{i}^{t-1} = \text{tanh}(\textbf{U}_q\textbf{h}_{i}^{t-1}+\textbf{b}_q),
\end{equation}
where $\textbf{U}_q$ and $\textbf{b}_q$ are model parameters. % that are used to  convert the hidden representation to transferred variables. 

Similar to LSTM, we generate a candidate cell state $\bar{\textbf{c}}_i^t$, a forget gate $f_i^t$ and a input gate $g_i^t$ by combining $\textbf{\textbf{x}}_i^t$ and the hidden representation at previous time step $\textbf{h}_i^{t-1}$, as follows: \begin{equation}
\small
\begin{aligned}
\bar{\textbf{c}}_i^t &= \text{tanh}(\textbf{U}_c \textbf{h}_i^{t-1} + \textbf{V}_c \textbf{x}_i^t+\textbf{b}_c),\\
\textbf{f}_i^t &= \sigma(\textbf{U}_f \textbf{h}_i^{t-1} + \textbf{V}_f \textbf{x}_i^t+\textbf{b}_f),\\
\textbf{g}_i^t &= \sigma(\textbf{U}_g \textbf{h}_i^{t-1} + \textbf{V}_g \textbf{x}_i^t+\textbf{b}_g),
\end{aligned}
\end{equation}
where $\{\textbf{U}_{l},\textbf{V}_l,\textbf{b}_l\}_{l=c,f,g}$ are model parameters.

% Then the LSTM generates a forget gate $f^t$, an input gate  $g^t$, and an output gate  via sigmoid function $\sigma(\cdot)$, as follows:
% %\vspace{-.07in}
% \begin{equation}
% \small
% \begin{aligned}
% %f^t = \sigma(W^f_h h^{t-1} + W^f_x x^t).
% \textbf{f}^t &= \sigma(\textbf{W}_f^h \textbf{h}^{t-1} + \textbf{W}_f^x \textbf{x}^t+\textbf{b}_f),\\
% \textbf{g}^t &= \sigma(\textbf{W}_g^h \textbf{h}^{t-1} + \textbf{W}_g^x \textbf{x}^t+\textbf{b}_g)
% \end{aligned}
% \end{equation}

After gathering the transferred variables for all the segments, we develop a new recurrent cell structure for each segment $i$ that integrates the transferred variables from its upstream segments into the computation of the cell state $\textbf{c}_i^t$. The forget gate $\textbf{f}_i^t$ and input gate $\textbf{g}_i^t$ are used to filter the information from previous time step and current time. This process can be expressed as: 
\begin{equation}
\small
    \textbf{c}_{i}^{t} = \textbf{f}_{i}^{t}\otimes (\textbf{c}_{i}^{t-1}+\sum_{(j,i)\in \mathcal{E}}\textbf{W}_{ji}\textbf{q}_{j}^{t-1})+\textbf{g}_{i}^t\otimes\bar{\textbf{c}}_{i}^t,
    \label{conv}
\end{equation}
where $\otimes$ denotes the entry-wise product.

%\yell{Mention the special case of head water. }
We can observe that the forget gate not only filters the previous information from the  segment $i$ itself but also from its neighbors (i.e., upstream segments). Each upstream segment $j$ is weighted by the adjacency level $\textbf{A}_{ji}$ between $j$ and $i$. When a river segment has no upstream segments (i.e., head water), the computation of %cell state in Eq.~\ref{conv} 
$\textbf{c}_i^t$ is the same as with the standard LSTM. In Eq.~\ref{conv}, we use $\textbf{q}_{j}^{t-1}$ %in Eq.~\ref{conv} 
%computed at 
from the previous time step %because there is usually 
because of the time delay in transferring the influence from upstream to downriver segments (the maximum travel time is approximately one day according to PRMS).

Then we generate the output gate $\textbf{o}^t_i$ to filter the cell state at $t$ and %. Then we compute the new cell state and 
output the hidden representation:% as follows: %and the hidden representation as: 
\begin{equation}
\small
\begin{aligned}
\textbf{o}_i^t &= \sigma(\textbf{U}_o \textbf{h}_i^{t-1} + \textbf{V}_o \textbf{x}_i^t+\textbf{b}_o),\\
\textbf{h}_i^t &= \textbf{o}_i^t\otimes \text{tanh}(\textbf{c}_i^t),
\end{aligned}
\label{hid}
\end{equation}
where $\{\textbf{U}_o,\textbf{V}_o,\textbf{b}_o\}$ are model parameters.

%After obtaining the cell state, we can compute the hidden representation $\textbf{h}_i^t$ by following Eq.~\ref{hid}. 

Finally, we generate the predicted output from the hidden representation as follows:
\begin{equation}
\small
    \hat{\textbf{y}}_i^t = \textbf{W}_y \textbf{h}_i^t+\textbf{b}_y,
    \label{prd}
\end{equation}
where $\textbf{W}_y$ and $\textbf{b}_y$ are model parameters.

%We apply the proposed model to a network of river segments. 
After applying this recurrent process to all the time steps, we define a loss using true observations %Given the set of true observations 
$\textbf{Y}=\{\textbf{y}_{i}^{t}\}$ that are collected at certain time steps and certain segments, as follows: % our training loss is defined as:
\begin{equation}
\small
    \mathcal{L}_{\text{RGrN}} = \frac{1}{|\textbf{Y}|} \sum_{\{(i,t)|\textbf{y}_{i}^{t}\in \textbf{Y}\}} (\textbf{y}_{i}^{t}-\hat{\textbf{y}}_{i}^{t})^2.
    \label{loss_PGRGrN}
\end{equation}

%The training of RGrN can be done using the standard back-propagation algorithm. 

\subsection{Markov Decision Process for Active Learning}

% motivation

% high-level description

The real-time labeling problem can be naturally formulated as a Markov sequential decision making problem. Here each state corresponds to  the current status of the predictive model given the input of currently observed samples. 
%After we take an action of labeling of not labeling observed data samples, the model will be updated and fed into new data at next time step, and thus returns  the next state. %Here the action represents a step in active learning, i.e., the selection of new query samples. 
% detailas
%Given the observed data at $t$ and current model status, a
Given the current state, an intelligent agent (i.e., the decision model) needs to take the optimal actions that maximize the potential reward. Here the actions represent the decisions of whether we will label each river segment at the current time $t$. The reward can be measured by the  improvement of predictive performance on an independent hold-out dataset. Our goal is to get a decision model that can maximize the gain of performance improvement.

%Below we will describe the details of the decision model. %We represent the state %by combining hidden embeddings, predictions and uncertainty computed by the predictive model 
%at the current time $t$. %More s
Specifically, we represent the state variable at each time $t$ as  $\textbf{S}^t\in \mathbb{R}^{N\times (H+3)}$,  %to 
%represent the state variable at each time $t$, 
where each row $\textbf{s}_i^t=[\textbf{h}_i^t, \hat{\textbf{y}}_i^t,\textbf{u}_i^t,\textbf{b}^t]$ is the concatenation of model embeddings $\textbf{h}_i^t$, prediction $\hat{\textbf{y}}_i^t$, and uncertainty $\textbf{u}_i^t$ of the segment $i$ obtained from the predictive model,  as well as the remaining budget $\textbf{b}^t$. Here the uncertainty level is estimated by Monte Carlo drop-out-based strategy used in previous work~\cite{daw2020physics,gal2016dropout}. Actions taken at each time are encoded by an $N$-by-$2$ matrix $\textbf{A}^t$ where each row $\textbf{a}^t_{i.}$ is a one-hot vector indicating whether we label the segment $i$ ([1, 0]) or not ([0, 1]).

The decision model is used to determine optimal actions to take at each time. In particular, the decision model takes the input of current state $\textbf{S}^t$ and outputs the potential future reward when the agent performs each possible action $\textbf{A}^t$
, i.e., the $Q$ value $Q(\textbf{S}^t,\textbf{A}^t)$. % for each selected action. Here the $Q$ value $Q(\textbf{S}^t,\textbf{A}^t)$ represents the overall expected reward when the agent is currently in state $\textbf{S}^t$ and performs action $\textbf{A}^t$, and continue taking actions using the same decision model. 
To train such a decision model, we need to estimate training labels for the $Q$ values by dynamic programming. 
Specifically, we consider a set of four-tuple samples over the sequence which consists of  $\{\textbf{S}^t,\textbf{A}^t, R(\textbf{S}^t,\textbf{A}^t), \textbf{S}^{t+1}\}$, where $R(\textbf{S}^t,\textbf{A}^t)$ is the immediate reward of performing actions $\textbf{A}^t$ at state $\textbf{S}^t$ and this can be measured as the reduction of prediction RMSE on an independent  hold-out dataset, and $\textbf{S}^{t+1}$ is the new model state after we observe data at $t+1$. 
We can estimate the training labels $Q(\textbf{S}^t,\textbf{A}^t)$ via dynamic programming, as follows:
\begin{equation}
    Q(\textbf{S}^t,\textbf{A}^t) = R(\textbf{S}^t,\textbf{A}^t)+\gamma \max_{\textbf{A}^{t+1}} Q(\textbf{S}^{t+1},\textbf{A}^{t+1}),
\end{equation}
where $\gamma$ is a discount factor. 
% where the immediate reward $R(\textbf{S}^t,\textbf{A}^t)$ is measured as the reduction of prediction RMSE on the hold-out dataset. 

During the prediction phase, we can select  optimal actions that maximize the $Q$ value, as follows:
\begin{equation}
    \textbf{A}^{t}_* = \text{argmax}_{\textbf{A}^t} Q(\textbf{S}^t,\textbf{A}^t)
\end{equation}

Given a large number of river segments, the action space will be exponentially large ($2^N$) and thus the model learning can be computationally intractable. In this work, we consider a simplified action space by performing actions independently on each river segment given its current state so we reduce the possible actions to be $2N$. Nevertheless, the selected actions for different river segments are still related to each other since we embed the spatial context of each segment in its state vector. 

% simplified action space:

% given state from one segment, determine its own action. They are not independent as the state vectors are collected a global model. 

% \vspace{-.1in}
\begin{algorithm}[!h]
\footnotesize
\caption{Training procedure of GR-REAL.}
\begin{algorithmic}[1]
\REQUIRE Training data $\textbf{X}_{tr} = \{\textbf{x}_i^t\}$ for $i=1:N$ and $t=1:T1$, and their available labels $\{\textbf{y}_i^t\}$; Holdout data $\textbf{X}_{hd} =\{\textbf{x}_i^t\}$ for $i=1:N$ and $t=T1+1:T$, and their available labels $\{\textbf{y}_i^t\}$;
% Test data $\textbf{X}_{te} =\{\textbf{x}_i^t\}$ for $i=1:N$ and $t=T:T+M$\\
%Budget $B$.
%\ENSURE Predictive model learned from collected labels on the test data.\\
\FOR{training epoch $k\leftarrow 1$ to $\text{Train\_iteration}$}
\STATE{Labeled set $\mathcal{L}=\{\}$, state transition set $\mathcal{M}$=\{\}.}
\STATE initialize the predictive model.
\FOR{time step $t\leftarrow 1$ to $T1$}
\IF{remaining budget$\le$0}
\STATE{Break}
\ENDIF
\STATE{Generate $\textbf{h}^t$, $\hat{\textbf{y}}^t$, $\textbf{u}^t$ %on training data 
by  the predictive model.}
\STATE{Concatenate the obtained values into the state vector $\textbf{s}^t$.}
\IF{$t>1$}
\STATE{Add  $(\textbf{S}^{t-1},\textbf{S}^{t},\textbf{A}^{t-1}, R(\textbf{S}^{t-1},\textbf{A}^{t-1}))$ to  $\mathcal{M}$.}
\STATE{Update the decision model using samples in $\mathcal{M}$.}
\ENDIF
\FOR{river segment $i\leftarrow 1$ to $N$}
\STATE{Predict $Q(\textbf{s}_i^t,\cdot)$ for different actions by the decision model.} 
\STATE{Select the action $\textbf{a}_i^t$ that leads to the highest $Q$ value. }
\ENDFOR % For j
\STATE{Update $\mathcal{L}$ with labeled samples and reduce the budget.}
\STATE{Update the predictive model using samples in $\mathcal{L}$ and measure the performance improvement on $\textbf{X}_{hd}$ as the reward $R(\textbf{S}^t,\textbf{A}^t)$.}
\ENDFOR % For i
\ENDFOR % For epoch
\end{algorithmic}
\label{algorithm}
\end{algorithm}

\vspace{-.15in}

\subsection{Implementation details}

We summarize the training process in Algorithm~\ref{algorithm}. Here we wish to clarify the difference between the training procedure and the test procedure. It is noteworthy that we repeatedly train our model by going through the training period for multiple passes (Line 1 in Algorithm~\ref{algorithm}). This is critical for refining the decision model given the fact that it can  make poor decisions (especially for the first few time steps) during the first few passes. When we apply the trained model to the test period (i.e., $T+1$ to $T+M$), we can only take one pass over the test data in a typical setting of real-time active learning. % and model update. 
In this case, the decision model cannot change any decisions that have been made in the past.

During the training process, we also allow the selection of random query samples from the training period with a probability of 0.5\%. This helps the decision model to explore a diverse set  of training samples and thus have a better estimate of the potential reward resulted from each labeling action.

\subsection{Policy Transfer}
Due to the substantial human effort and material costs to collect observation data, we often have limited data for training ML models. For example, the Delaware River Basin is one of the most widely studied basins in the United States but still has  less than 2\% daily temperature observation for over 80\% of its internal river segments~\cite{read2017water}. Such sparse data makes it challenging to train %the predictive model (i.e., the RGrN model) and 
a good decision model. % given their complexity. 
%Given their complexity, the RGrN model and the decision model trained with limited observed data can lead to poor performance.   

%If physical knowledge can be used to help inform the initialization of the weights, model
%training can accelerated (i.e., require fewer epochs for training) and also need fewer training samples to achieve good performance. 

To address this issue, we propose to transfer the model learned from simulation data produced by physics-based models. Physics-based models are built based on known physical relationships that transform input features to the target variable. In the context of modeling river networks, we use PRMS-SNTemp model to simulate target variables (i.e., temperature and streamflow) given the input drivers. Hence, for the input features $\textbf{x}_{i:N}^{1:T}$ for all the $N$ river segments in a past period with $T$, we can generate their simulated target variables  $\textbf{y}_{i:N}^{1:T}$ by running the PRMS-SNTemp model.   
% for each segment $i$ and for  given the input drivers $\textbf{x}_{i}^{1}, \textbf{x}_{i}^2, ..., \textbf{x}_{i}^T$. 
We can pre-train  the decision model using the simulation data. Then we can refine the decision model when it is  applied to true observations.

It is noteworthy that simulation data from the PRMS-SNTemp model are imperfect but they only provide a synthetic realization of physical responses of a river systems to a given set of input features. Nevertheless, pre-training a neural network using simulation data allows the network to emulate a synthetic but physically realistic phenomena. %This process results in a more accurate and physically consistent initialized status for the learning model. 
When applying the pre-trained model to a real system, we fine-tune the model using true observations. Our hypothesis is that the pre-trained model is much closer to the optimal solution and thus requires less true labeled data to train a good quality model. %In our experiments, we show that such pre-trained models can achieve high accuracy given only a few observed data points. 

\section{Experimental Results}
\subsection{Dataset and test settings}
\label{sec:dataset}

The dataset is pulled from U.S. Geological Survey's National Water Information System~\cite{us2016national} and  the Water Quality Portal~\cite{read2017water}, the largest standardized water quality data set for inland and coastal waterbodies~\cite{read2017water}. %Streamflow and stream temperature observations are pulled from the these databases and matched to our geospatial modeling fabric. 
Observations at a specific latitude and longitude were matched to river segments that vary in length from 48 to 23,120 meters. The river segments were defined by the national geospatial fabric used for the National Hydrologic Model as described by Regan et al.~\cite{regan2018description}, and the river segments are split up to have roughly a one day water travel time. We match observations to river segments by snapping observations to the nearest river segment within a tolerance of 250 meters. Observations farther than 5,000 m along the river channel to the outlet of a segment were omitted from our dataset. %Segments with multiple observation sites were aggregated to a single mean daily streamflow or water temperature value.    

%\yell{Jake: please add a brief description about how this data is collected and relevant data source citations.}

%In particular, w
We study a subset of the Delaware River Basin with 42 river segments that feed into the mainstream Delaware River at Wilmington, DE. We use input features at the daily scale from Oct 01, 1980 to Sep 30, 2016 (13,149 dates). The input features have 10 dimensions which include daily average precipitation, daily average air temperature, date of the year, solar radiation, shade fraction, potential evapotranspiration and  the geometric features of each segment (e.g., elevation, length, slope and width). %Air temperature and precipitation values were derived from the Daymet gridded dataset. Other input features (e.g., shade fraction, solar radiation, potential evapotranspiration) are difficult to  measure frequently, and we use values produced by the PRMS-SNTemp model as its internal variables. 
Water temperature observations were available for 32 segments but the temperature was observed only on certain dates. The number of temperature observations  available 
for each observed segment ranges from 1 to 9,810 with a total of  51,103 %temperature data observed on different dates and different segments
observations across all dates and segments. Streamflow observations were available for 18 segments. The number of  streamflow observations  available 
for each  observed segment ranges from 4,877 to 13,149 %and in total we have 
with a total of 206,920 observations across all dates and segments. %streamflow data points observed on different dates and different segments. 

% \yell{how to generate adjacency matrix?}

We divide the available data into four periods, training period (Oct 01, 1980 - Sep 30, 1989), hold-out period (Oct 01, 1989 - Sep 30, 1998), test period (Oct 01, 1998 - Sep 30, 2007) and evaluation period (Oct 01, 2007 - Sep 30, 2016). We assume the current time $T+1$ starts from Oct 01, 1998, i.e., the start of the testing period. We will apply the trained decision model to the test period to collect new samples and use them to train the predictive model. Then the predictive model will be evaluated in the evaluation period.

%Since traditional active learning methods cannot be directly applied to our task, w
We will compare with a set of baselines: %customized baselines by modifying some traditional baselines:

\begin{itemize}
\vspace{-.1in}
\itemsep0em 
    \item Random selection: This method  assumes the access to the entire set of testing data (so it is a pool-based method). % at the beginning. The w
    We randomly select a subset of query samples (size equal to the budget) from the test data for labeling. 
    \item Uncertainty: For each time step, we estimate the uncertainty of each data sample (i.e., each node in the graph) using the method presented in previous work~\cite{daw2020physics,gal2016dropout}. Then we select the query samples if their uncertainty values are above a threshold. Such threshold is estimated in the training period.   
    \item Uncentainty+Centrality+Density (UDC): This is a graph-based active learning method. For each time step, we compute the weighted summation of uncertainty (measured by~\cite{daw2020physics,gal2016dropout}), the centrality and the density of each node following the previous work~\cite{cai2017active,gao2018active}. Then we select the query samples if their summation values are above a threshold. Such threshold is estimated in the training period. 
    \item ROAL~\cite{huang2019improvement}: This methods solves the real-time labeling problem using reinforcement learning. It considers each segment separately but uses an LSTM to estimate the Q value.  
    \vspace{-.1in}
\end{itemize}

Here the first three baselines use the same predictive model as our proposed GR-REAL method, %. The GR-REAL method 
which uses a one-layer graph structure. Since the adjacency matrix $\textbf{W}$ includes $(i,j)$ pairs for segment $j$ anywhere downstream of segment $i$, the one-layer graph structure can still capture spatial dependencies between river segments that are not directly connected.  
We generate the adjacency matrix $\textbf{W}$ based on the stream distance between each pair of river segment outlets, represented as $\text{dist}(i,j)$. We standardize the stream distance  and then compute the adjacency level as $\textbf{W}_{ij}=1/(1+\text{exp}(\text{dist}(i,j)))$ for each edge $(i,j)\in\mathcal{E}$.  The hidden variables in the networks have a dimension of 20. The hyper-parameter $\gamma$ is set as 0.8. For uncertainty estimation, we have used a dropout probability of 0.2 and randomly created 10 different dropout networks. During the training and testing, we also set a yearly limit so as to avoid having too many query samples selected from a single year. Here we set the yearly limit to be $1.2\times\text{Budget}/\#\text{years}$

% \textcolor{red}{To make sure data are equally distributed over entire test period, yearly budget, 1.2}

\subsection{Predictive performance}
\begin{table}[!t]
\footnotesize
\newcommand{\tabincell}[2]{\begin{tabular}{@{}#1@{}}#2\end{tabular}}
\newcolumntype{C}[1]{>{\centering\arraybackslash}p{#1}}
\centering
\caption{RMSE (standard deviation) for streamflow modeling ($m^3/s$) with different budgets. }
\begin{tabular}{l|C{0.7cm} C{0.7cm} C{0.7cm} C{0.7cm} C{0.7cm} C{0.7cm}}
\hline
\textbf{Samples} &\textbf{100} & \textbf{300} & \textbf{500} & \textbf{1000}& \textbf{2000}\\ \hline 
\multirow{2}{*}{Random} & 6.24 & 5.97 & 5.92   &5.66 &5.59 \\ 
& (0.52) & (0.34) & (0.36)  & (0.20)&(0.23)\\ \hline
\multirow{2}{*}{Uncertain} & 6.16 &5.88 &5.83  &5.64 &5.43 \\
& (0.26) & (0.24) & (0.20)  & (0.12)&(0.14)\\ \hline
\multirow{2}{*}{UDC} & 6.12 &5.84 &5.76  &5.49 &5.44 \\
& (0.23) & (0.26) & (0.19)  & (0.12)&(0.12)\\ \hline
\multirow{2}{*}{ROAL} & 5.79 &5.74 &5.70  &5.22 &5.36 \\
& (0.25) & (0.22) & (0.13)  & (0.11)&(0.11)\\ \hline
\multirow{2}{*}{GR-REAL} & 5.33  &5.40 & 5.41  & 4.80 & 4.78 \\
& (0.21) & (0.22) & (0.14)  & (0.11)&(0.08)\\ 
\hline
\end{tabular}
\label{perf_flow}
\end{table}

\begin{table}[!t]
\footnotesize
\newcommand{\tabincell}[2]{\begin{tabular}{@{}#1@{}}#2\end{tabular}}
\newcolumntype{C}[1]{>{\centering\arraybackslash}p{#1}}
\centering
\caption{RMSE (standard deviation) for temperature modeling ($^\circ C$) with different budgets. }
\begin{tabular}{l|C{0.7cm} C{0.7cm} C{0.7cm} C{0.7cm} C{0.7cm} C{0.7cm}}
\hline
\textbf{Samples} &\textbf{100} & \textbf{300} & \textbf{500} & \textbf{1000}& \textbf{2000}\\ \hline
\multirow{2}{*}{Random} & 4.42 &3.89&  3.42&  3.31&  3.25\\ 
& (0.45)& (0.46)& (0.34)& (0.14)& (0.09) \\\hline
\multirow{2}{*}{Uncertain} & 4.16& 3.51& 3.40&   3.28 &3.27 \\
& (0.38)& (0.35)& (0.35)& (0.17)& (0.06) \\ \hline
\multirow{2}{*}{UDC} & 4.17& 3.56& 3.33& 3.24& 3.26 \\
& (0.29)& (0.24)& (0.26)& (0.12)& (0.04) \\ \hline
\multirow{2}{*}{ROAL} & 4.04& 3.54& 3.38& 3.24& 3.24 \\
& (0.27)& (0.19)& (0.19)& (0.13)& (0.08) \\ \hline
\multirow{2}{*}{GR-REAL} & 3.40&  3.34&  3.29&  3.23&  3.24 \\
& (0.18)& (0.16)& (0.15)& (0.09)& (0.05) \\ 
\hline
\end{tabular}
\label{perf_temp}
\end{table}

In Tables~\ref{perf_flow} and~\ref{perf_temp}, we report the performance of each method with different budget limits (i.e., the number of allowed labeled samples). We repeat each method five times and report mean and standard deviation of prediction RMSE (in the evaluation period). In general, the proposed GR-REAL outperforms other methods by a considerable margin. For temperature modeling, all the methods have similar performance when they have access to sufficient labeled samples (e.g., 2000 samples), but GR-REAL outperforms other methods when we have limited budget. Here the method based on model uncertainty, node density and centrality  can be less helpful because these measures indicate only the representativeness of samples at the current time but not the potential future benefit in a real-time sequence. The ROAL method also does not perform as well as the proposed method as it does not take into account the spatial context when estimating the potential reward of labeling each sample.

\begin{figure} [!h]
\centering
%\vspace{-.2in}
\includegraphics[width=0.5\columnwidth]{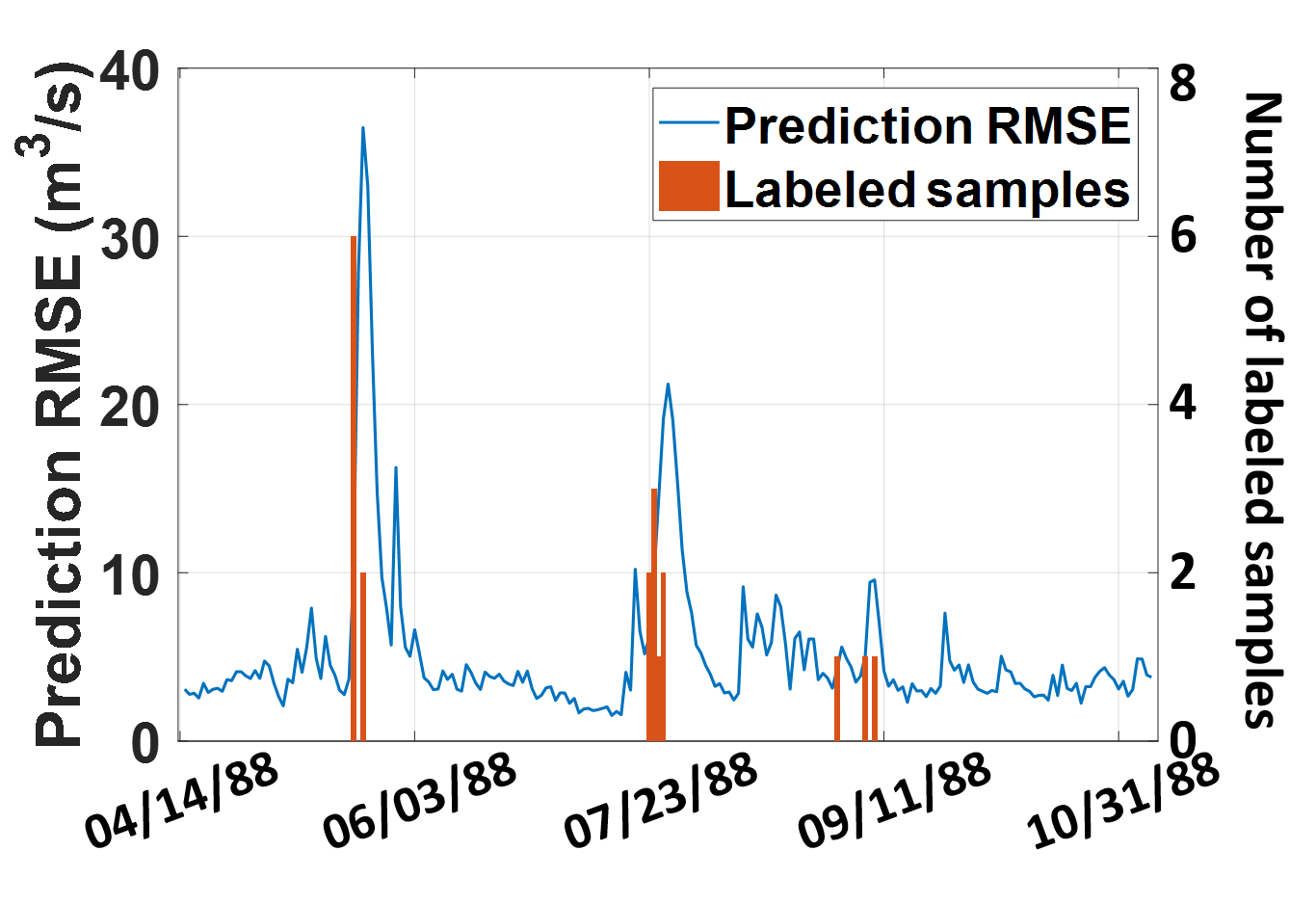}
\vspace{-.2in}
\caption{The relationship between the number of selected query samples and the change of prediction RMSE over time.}
\label{fig:track}
\end{figure}

In Fig.~\ref{fig:track}, we show an  example for the selected samples by GR-REAL in streamflow modeling (training period). It can be clearly seen that the many samples are labeled right at the time when the predictive model starts to get an increase of prediction error. Then the error decreases after we refine the model  using the labeled samples. This shows the effectiveness of the proposed method in automatically detecting  query samples which can bring large potential benefit for model training.

\subsection{Distribution of selected data}

\begin{figure} [!t]
\centering
%\raggedleft
\subfigure[]{ \label{fig:a}
\includegraphics[width=0.45\columnwidth]{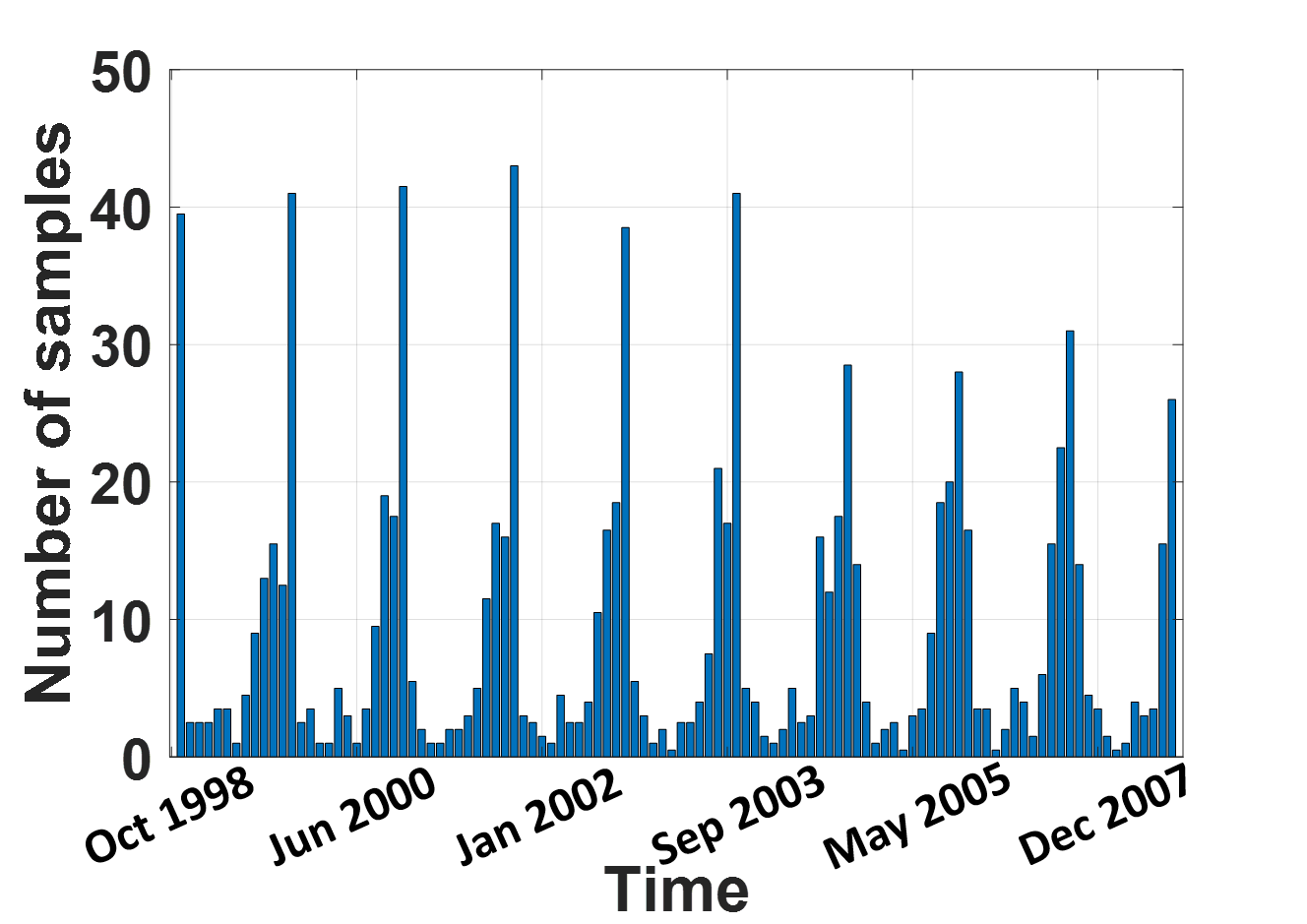}
}\hspace{-.1in}
\subfigure[]{ \label{fig:a}
\includegraphics[width=0.45\columnwidth]{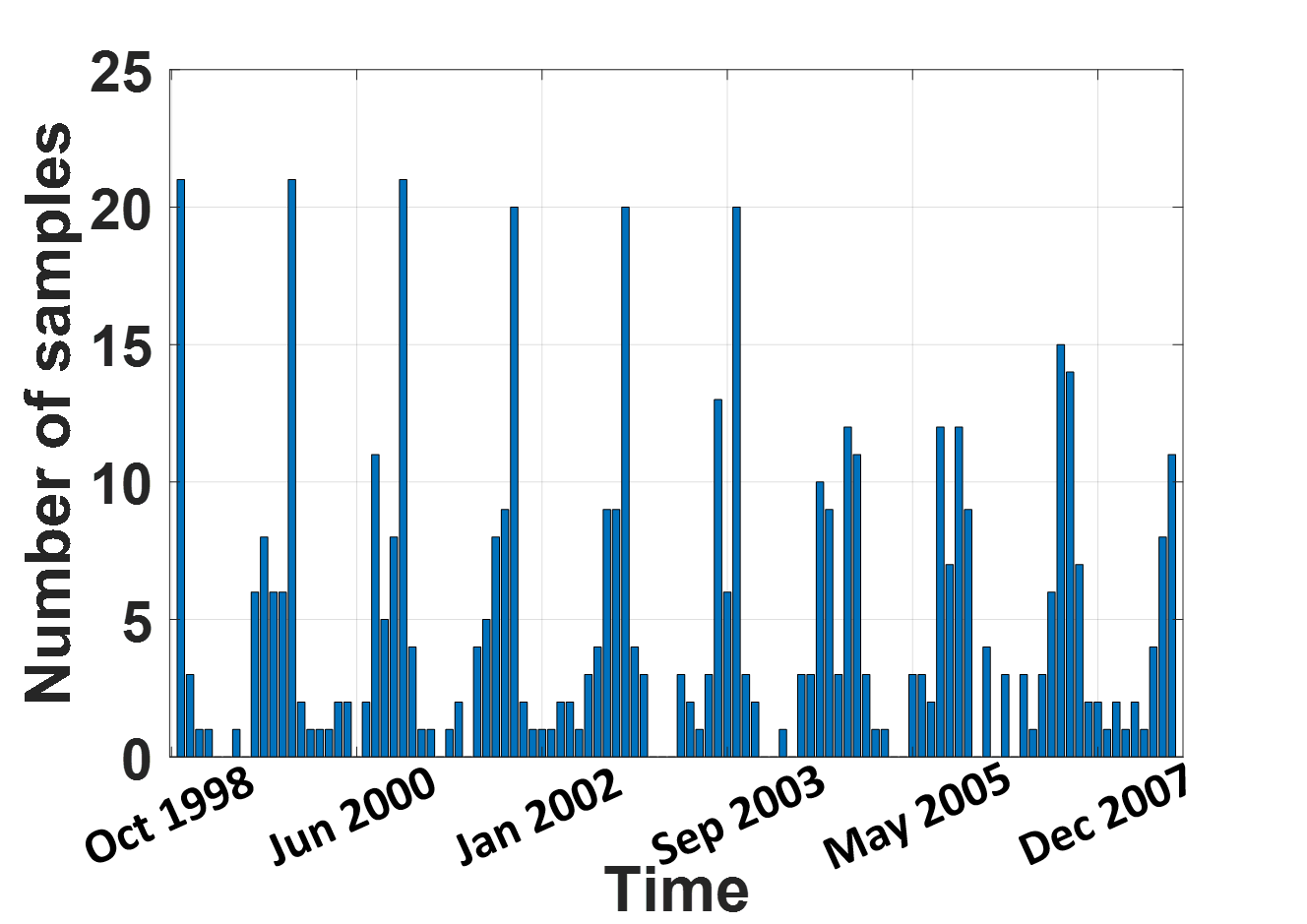}
}
\vspace{-.2in}
\caption{The distribution of selected query sample over time for temperature modeling given the budget of (a) 1000 and (b) 500.}
\label{fig:data_sel}
%\vspace{-.15in}
\end{figure}

In Fig.~\ref{fig:data_sel}, we show the distribution of selected data samples over the entire test period for temperature prediction. Given the budget limit of 1000, we observe that GR-REAL selects most samples in the summer period (Fig.~\ref{fig:data_sel} (a)), especially before 2004. This is because river temperature in the Delaware River Basin are usually %have a much stronger temperature variability 
much more variable in summer period and thus more samples are selected to learn these complex dynamic patterns. Since 2004, the model selects fewer data on the same ``peak" period since the potential reward for training the model decreases after it has learned from samples in the same period from previous years.  When we reduce the budget to 500, GR-REAL also selects more samples in other time periods than the summer period (especially after 2004) because the budget is limited and the model needs to balance the performance over the entire year to get the optimal overall performance.

%Compared with GR-REAL, we can see that the selected samples by \textit{Uncertainty} and \textit{UDC} are distributed in different periods. In these methods, a threshold is used to determine whether the model will select samples at each time. However, they do not dynamically update the threshold after the model is refined using the collected samples. Hence, these models cannot select most helpful samples in a real-time sequence.  

% temporal

\begin{figure} [!t]
\centering
%\raggedleft
\subfigure[]{ \label{fig:a}
\includegraphics[width=0.45\columnwidth]{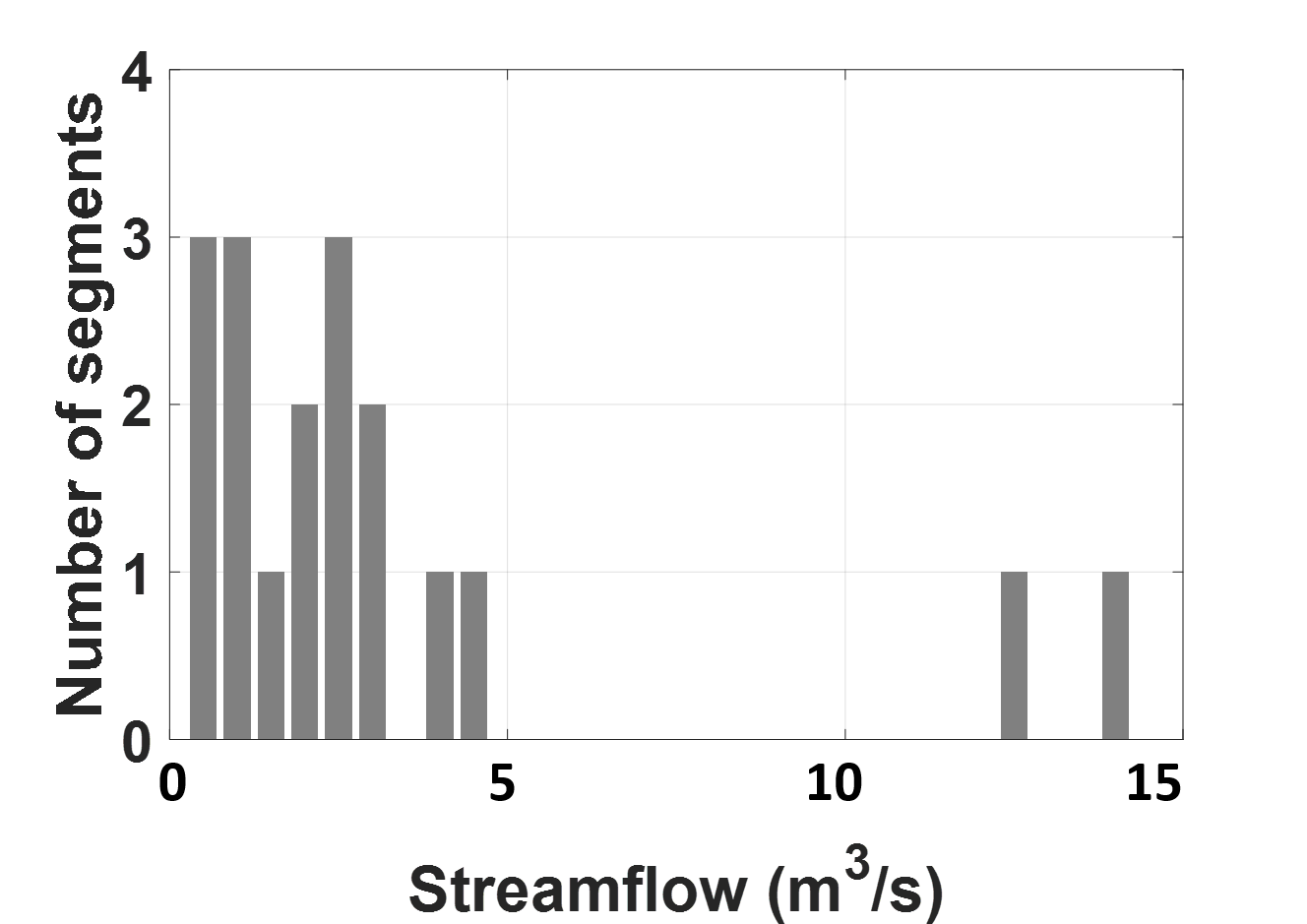}
}\hspace{-.1in}
\subfigure[]{ \label{fig:b}
\includegraphics[width=0.45\columnwidth]{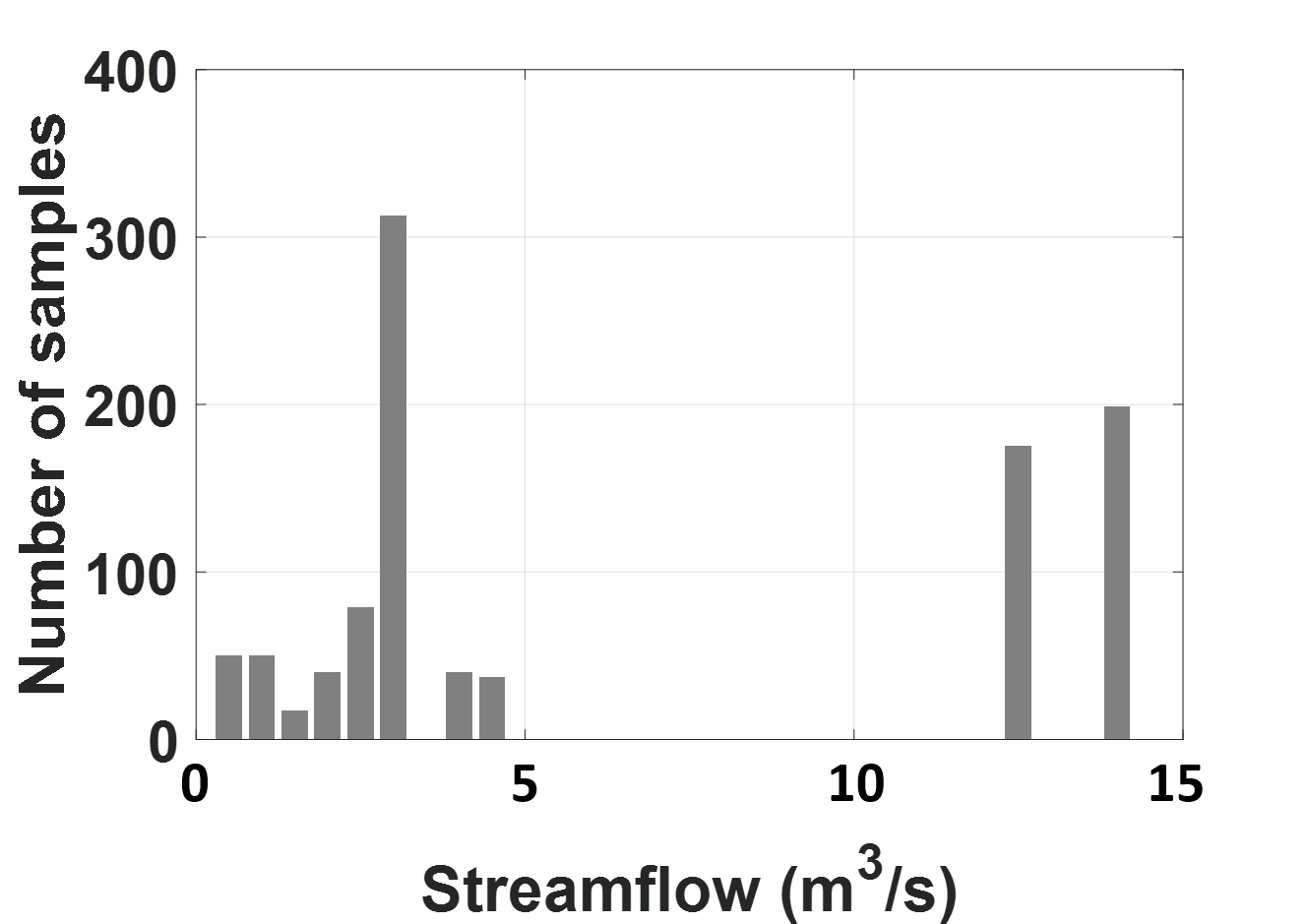}
}\vspace{-.2in}
\caption{(a) Number  of river segments  for each streamflow range. (b) Distribution of selected samples over river segments of different streamflow ranges. }
\label{fig:sel_sp}
%\vspace{-.15in}
\end{figure}
% % temporal
% \begin{figure*} [!h]
% \centering
% %\raggedleft
% \subfigure[]{ \label{fig:a}
% \includegraphics[width=0.42\columnwidth]{AQL_1000.png}
% }
% \subfigure[]{ \label{fig:b}
% \includegraphics[width=0.42\columnwidth]{uncertain_1000.png}
% }
% \subfigure[]{ \label{fig:b}
% \includegraphics[width=0.42\columnwidth]{cluster_1000.png}
% }
% \subfigure[]{ \label{fig:b}
% \includegraphics[width=0.42\columnwidth]{rand_1000.png}
% }
% \subfigure[]{ \label{fig:a}
% \includegraphics[width=0.42\columnwidth]{AQL_500.png}
% }
% \subfigure[]{ \label{fig:b}
% \includegraphics[width=0.42\columnwidth]{uncertain_500.png}
% }
% \subfigure[]{ \label{fig:b}
% \includegraphics[width=0.42\columnwidth]{cluster_500.png}
% }
% \subfigure[]{ \label{fig:b}
% \includegraphics[width=0.42\columnwidth]{rand_500.png}
% }
% \caption{(a) flow (b) temperature.}
% \label{fig:data_sel}
% \end{figure*}

% spatial

For the streamflow modeling, we show the distribution of selected samples over segments with different streamflow ranges. For each segment, we first compute its average streamflow value and we show the histogram of average streamflow over all the 18 segments (which have streamflow observations) in Fig.~\ref{fig:sel_sp} (a). Then we show in Fig.~\ref{fig:sel_sp} (b) the distribution of selected samples with respect to the average streamflow of the segments from which the samples are taken. We can see that most samples are selected from the two segments with largest streamflow values and from the segments with average streamflow around 3$m^3/s$. %The reason behind is 
We hypothesize that labeled samples from segments with highstreamflow ($>12m^3/s$) and middle flow range($1-5m^3/s$) are more helpful to refine the model so as to reduce the overall prediction error.  In contrast, more samples on low-flow range ($<1m^3/s$) can bring limited improvement to the overall error because errors made on low-flow segments tend to be much smaller. %even they can help ML models to better capture low-flow patterns. 
This is one limitation of the proposed method as accurate prediction on low-flow segments is also important for understanding the aquatic ecosystem. %We will keep this as our 
These limitations may be potential opportunities for future work.

% temporal

% \subsection{When the new samples are selected over the change of hold out/testing errors}

% temporal

\subsection{Performance of different graph models}

Here we show how the representation of graphs impacts the learning performance (Table.~\ref{fig:perf_graph}). We consider the following three representation: 1) Downstream-neighbor graph which is the graph used in our implementation. Here we add edges from segment $i$ to segment $j$ if segment $j$ is anywhere downstream of segment $i$. Consider three segments $\{a,b,c\}$ in a upstream-to-downstream consecutive sequence. We will include the edge $ab$, $bc$ and $ac$ in our graph and their adjacency levels are determined by the stream distance. 2) Direct-neighbor graph which only includes connected neighbors. In the above example, we will only create edges of $ab$ and $bc$.  3) No-neighbor graph which is equivalent to an RNN model trained using data from all the segments.  
The choice of graph representation can affect the selection of representative samples. The improvement from no-neighbor graph to direct-neighbor graph shows that the modeling of spatial context can help select more informative samples. The downstream-neighbor graph results in better performance since a river segment can impact downriver segments that are multi-hops away and thus incorporating multi-hop relationships can help  better embed the contextual information.

\begin{table}[!t]
\footnotesize
\newcommand{\tabincell}[2]{\begin{tabular}{@{}#1@{}}#2\end{tabular}}
\centering
\caption{The prediction RMSE (budget=500) using different graph structures. }
\begin{tabular}{|l|cc|}
\hline
\textbf{Graph} &streamflow($m^3/s$) & temperature($^\circ C$) \\ \hline 
Downstream neighbor & 5.41 & 3.29\\ 
Direct neighbor & 5.57& 3.33\\
No neighbor & 5.71 & 3.36\\
\hline
\end{tabular}
\label{fig:perf_graph}
%\vspace{-.1in}
\end{table}
\begin{figure} [!t]
\centering
%\raggedleft
\subfigure[]{ \label{fig:a}
\includegraphics[width=0.4\columnwidth]{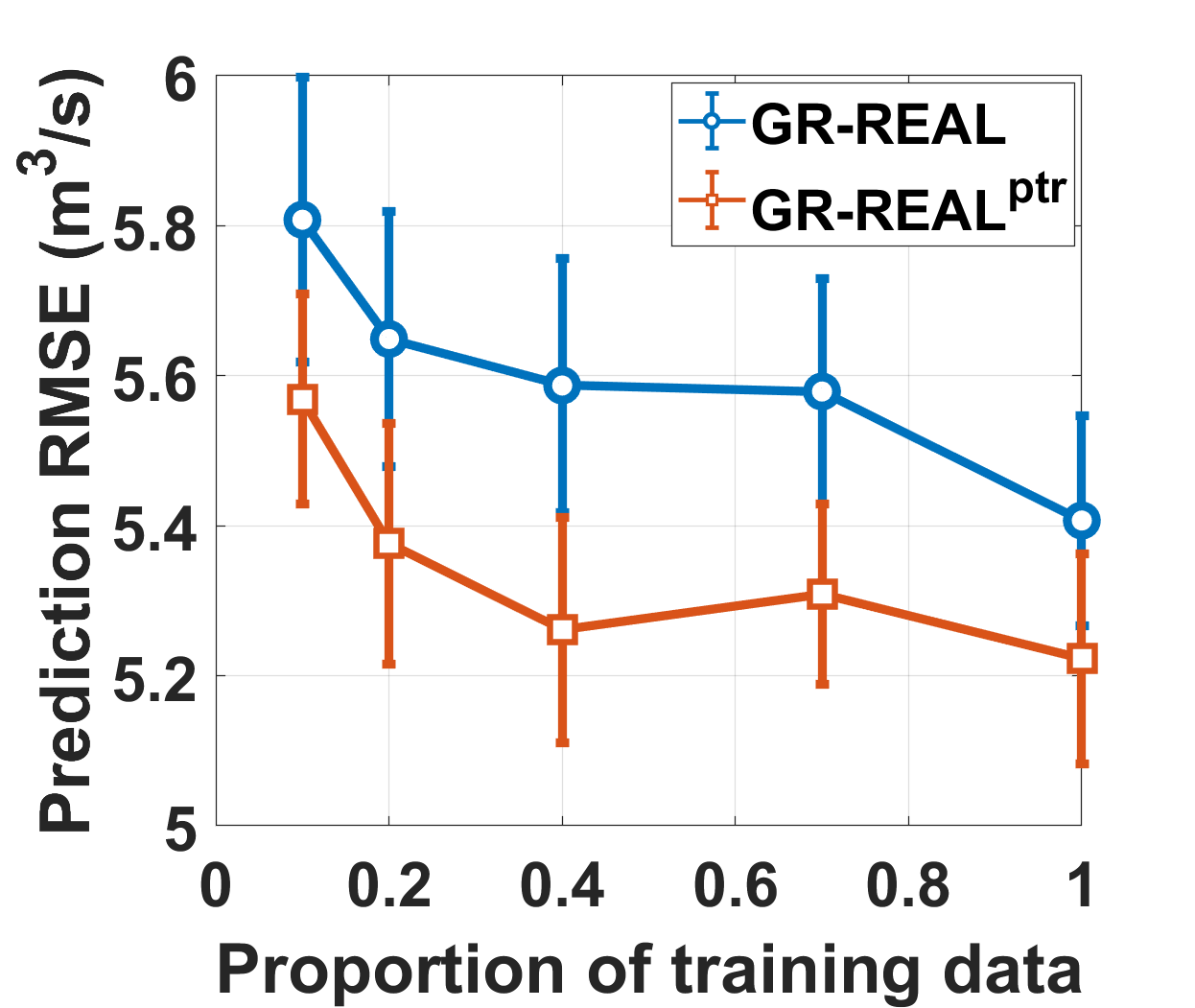}
}
\subfigure[]{ \label{fig:b}
\includegraphics[width=0.4\columnwidth]{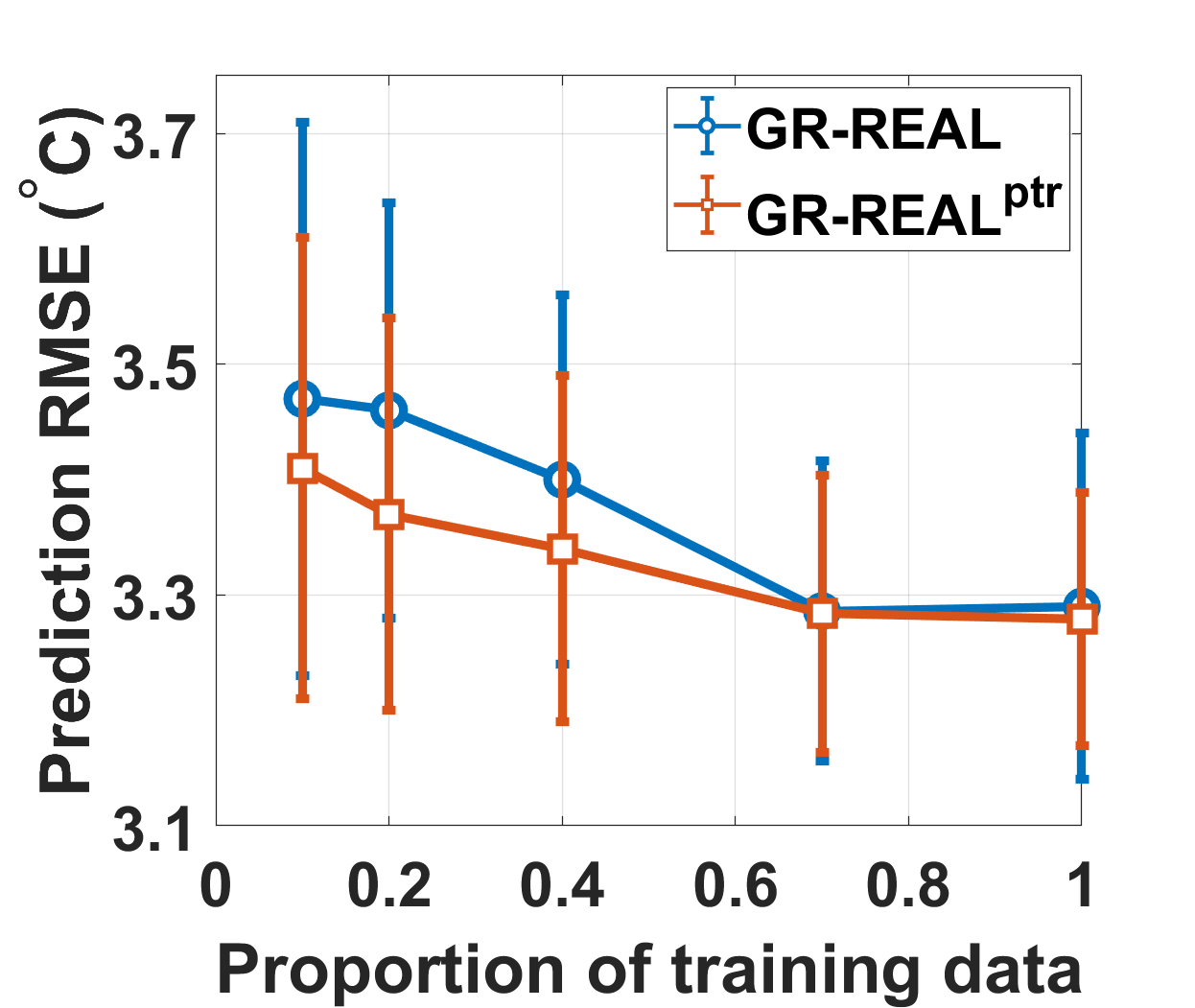}
}
\vspace{-.15in}
\caption{Prediction performance of GR-REAL before and after using the policy transfer in modeling (a) streamflow and (b) water temperature.}
\label{fig:pretrain}
%\vspace{-.15in}
\end{figure}
% all the downriver segments

% only the direct downriver segments

% no connections

\subsection{Policy transfer}
We also study the efficacy of policy transfer from simulation data when we have access to limited training data.
In particular, we randomly reduce the labeled samples for temperature and streamflow in the training period so that GR-REAL will learn different decision models. Then we will evaluate the performance using these decision models as reported in Fig.~\ref{fig:pretrain}. We can observe that the predictive performance decreases as we reduce the number of training data. This is because we can only learn a sub-optimal decision model  using limited training data and thus it may not be able to select the most helpful samples for training the predictive model. However, the method with policy transfer (GR-REAL$^\text{ptr}$) can produce better performance since the model is initialized to be much closer to its optimal state. In temperature prediction, we notice that GR-REAL and GR-REAL$^\text{ptr}$ have similar performance when using more than 70\% training data. This is because the training data is sufficient to learn an accurate decision model.

% \vspace{-.2in}

\section{Conclusion}
In this paper, we propose the GR-REAL framework which uses the spatial and temporal contextual information to select query samples in real time. %The proposed framework can be formulated as a reinforcement learning problem. 
We demonstrate the effectiveness of GR-REAL in selecting informative samples for modeling streamflow and water temperature in the Delaware River Basin. We also show that policy transfer can further improve the  performance when we have less training data.  The proposed method may also be used to
%these parameters to have a decision model that identified new sites at which to begin long-term monitoring. However, there are many 
measure other water quality parameters for which sensors are costly or too difficult to maintain (e.g., metal, nutrients, or algal biomass).

%in a wide range of scientific problems in climate science, hydrology, and agronomy to assist in data collection and field study.

% limitation
While GR-REAL achieves better predictive performance, it estimates potential reward based on accuracy improvement and thus remains limited in  selecting samples that indeed help understand an ecosystem. For example, the GR-REAL ignores low-flow segments as they contribute less to the overall accuracy loss. Studying this element of GR-REAL has promise for future work.

\section{Acknowledgments}
Any use of trade, firm, or product names is for descriptive purposes only and does not imply endorsement by the U.S. Government.

\bibliographystyle{plain}
\vspace{-.05in}
\bibliography{reference}
\end{document}